\documentclass[runningheads]{llncs}
\usepackage{graphicx}
\usepackage{amsmath}
\usepackage[colorlinks,citecolor=red,urlcolor=blue,bookmarks=false,hypertexnames=true]{hyperref}
\usepackage[super]{nth}
\usepackage[english]{babel}
\usepackage[autostyle]{csquotes}
\usepackage{siunitx}
\usepackage{setspace}
\usepackage{academicons}
\usepackage{xcolor}
\usepackage{svg}

\usepackage{amsmath}
\usepackage{amssymb}
\usepackage{amsfonts}
\usepackage{mathrsfs}
\usepackage{caption}
\usepackage{subcaption}

\usepackage{xspace}

\newcommand{\ie}{\textit{i.e.}\xspace}

\newcommand{\eg}{\textit{e.g.}\xspace}
\newcommand{\etc}{\textit{etc.}\xspace}

\DeclareGraphicsExtensions{.pdf,.png,.jpg,.bmp}

\usepackage{subcaption}
\usepackage{listings}

\begin{document}
\title{Do intermediate feature coalitions aid explainability of black-box models?}
%
%

\author{Anonymous}
\author{Minal Suresh Patil\orcidID{0000-0003-0026-5503} \and
Kary Främling\orcidID{0000-0002-8078-5172}}
%

%


\institute{Umeå universitet \newline
\email{\{minalsp,kary.framling\}@cs.umu.se}}

\maketitle              
\begin{abstract}
This work introduces the notion of intermediate concepts based on levels structure to aid explainability for black-box models. The levels structure is a hierarchical structure in which each level corresponds to features of a dataset (i.e., a player-set partition). The level of coarseness increases from the trivial set, which only comprises singletons, to the set, which only contains the grand coalition. In addition, it is possible to establish meronomies, i.e., part-whole relationships, via a domain expert that can be utilised to generate explanations at an abstract level. We illustrate the usability of this approach in a real-world car model example and the Titanic dataset, where intermediate concepts aid in explainability at different levels of abstraction.
\end{abstract}

\section{Introduction}
A mathematical theory of coalition behavior is the theory of n-person cooperative games. Determining which of the potential coalitions may be anticipated to form and what will be the ultimate distribution of payoffs to each player is a fundamental problem in game theory. In the field of coalition games, research on coalition formation has consistently attracted attention. Often, it is implicitly assumed that any coalition may form and that all actors will share in the value of the grand coalition. It is possible that in some situations the players would want to organize themselves in another way because the coalition building issues could be influenced by a variety of variables (such as the sharing rule in use). Based on the factors that influence the nature of coalitions, the research on coalition formation in cooperative games can be divided into two categories.

First,it concentrates on cooperative games when player collaboration is constrained by predetermined social, technological, or other structures. The coalition structure was put forth by Aumann and Dreze~\cite{aumann1974cooperative} as a division of players into coalitions that are unrelated to one another. Negotiations are only permitted until the coalition structure has been established inside each coalition that makes up the structure. By presuming that eventually the grand coalition will form with the coalitions that make up the coalition structure as members, Owen elaborates on the coalition structure~\cite{owen1977values}. Cooperative games with communication structures were first introduced by Myerson~\cite{myerson1980conference} where an undirected graph that represents players as nodes and enables for collaboration only amongst coalitions that are connected by the graph. In games with precedence constraints, where the set of players is organized according to some precedence relation, Faigle and Kern~\cite{faigle1992shapley} considered only coalitions that aim to satisfy the constraints in the sense that if a player is a member of a coalition, all players who come before that player must also be members of the same coalition. Hyper-graph communication scenarios, in which the nodes are the players and the links are the subgroups of players who can communicate, were first described by Van den Nouweland et al.~\cite{van1992allocation}. The assumption is that communication can only exist within a conference and that for communication to happen, all players must be present. A paradigm for cooperative games with levels structure was first presented by Winter~\cite{winter1989value}. A succession of coalition structures known as \emph{levels structures} are generated from each other by aggregating coalitions each of which encompasses the numerous coalition cooperations of the prior coalition structure. \\
Second, a specific solution concept or a group of characteristic functions are utilized to construct the coalition. The notion of coalition formation motivated by a coalition structure was covered by Hart and Kurz~\cite{hart1996bargaining}. In order to choose some coalition structures from a strategy profile that is a strong equilibrium~\cite{aumarm2016acceptable}, meaning,  there is no straightforward payoff incentive for players to modify this structure, they first established a valuation criterion using the Owen value~\cite{owen1977values} for each individual player in a given coalition structure. A specific number of players who can provide specific payoffs via cooperation are termed a transferable utility cooperative game (TU). There are no restrictions on player cooperation; it is defined by the value that every coalition may achieve on its own. However, this setup is not feasible in all situations where cooperation is restricted. In order to model this, TU games were proposed by Aumann and Dreze\cite{aumann1974cooperative} where the grand coalition was divided as disjoint prior unions and each of these unions were considered a whole without any interactions. However, Owen~\cite{owen1977values} relaxed this assumption and allowed players within a prior union to interact with other unions. The same rationale that compels players to unionize to boost their bargaining power during the distribution of the grand coalition's value should also motivate unions to form larger unions. The process can continue until the grand coalition emerges which leads to the formation of hierarchical or levels structure~\cite{winter1989value}. \\ An example is an organisation with various levels of a structure while distributing resources at all levels. In a university, at the top-level faculties are grouped by the university and a level, we have multiple departments grouped by faculties, and lastly, lecturers are grouped by departments. TU games with a levels structure are appealing due to \emph{level structure's} strong descriptive power. In this work, we focus on how intermediate concepts in machine learning can be represented as \emph{levels structure} for explaining machine learning models.
\section{Technical Preliminaries}
\subsection{Coalition Games}

A pair for a cooperative game is denoted as $(N, v)$, where $N=\{1,2, \ldots, n\}$ represents the players and $v: 2^N \rightarrow \mathbb{R}_{+}$ with $v(\emptyset)=0$ as the characteristic function.
A subset $S \subseteq N$ is a \emph{coalition} and $v(S)$ is the worth of $S$. For convenience, we write $\{i\}$ as $i$. If there is no ambiguity, we identify the game $(N, v)$ with its characteristic function $v$. The set of all cooperative games over $N$ is denoted by $G^N$. The restriction of a game $(N, v)$ to a coalition $S \subseteq N$, denoted by $\left(S, v_S\right)$, is defined by $v_S(T)=v(S \cap T)$ for all $T \subseteq S$. We represent $(S, v)$ instead of $\left(S, v_S\right)$ and $|\cdot|$ represents the number of players in a set.
\begin{itemize}
    \item A \emph{cooperative game} $(N, v)$ is said to be zero-monotonic if $v(S \cup i) \geq v(S)+v(i)$ for any $S \subseteq N \backslash i$.
    
    \item A \emph{cooperative game} $(N, v)$ is said to be monotonic if $v(S) \geq v(T)$ for all $S, T \subseteq N$, $S \supseteq T$
    
    \item A \emph{cooperative game} is superadditive if for all $S, T \subseteq N$ with $S \cap T=\emptyset$, it holds that $v(S \cup T) \geq v(S)+v(T)$ and $S G^N$ represents the set of super-additive cooperative games.
    
    \item A \emph{cooperative game} $v$ is convex if $v(S \cup T)+v(S \cap T) \geq v(S)+v(T)$ for all $S, T \subseteq N$
\end{itemize}

\begin{definition}
For any $\emptyset \neq T \subseteq N,\left(N, u_T\right) \in G^N$ is called a unanimity game, where for any $S \subseteq N$,

$$
u_T(S)= \begin{cases}1 & \text { if } T \subseteq S \\ 0 & \text { otherwise. }\end{cases}
$$
\end{definition}

It is well established for any unanimity games $u_T, \emptyset \neq T \subseteq N$, form a basis for $G^N$, i.e., each game $(N, v) \in G^N$ can be expressed by $v=\sum_{T \subseteq N, T \neq \emptyset} \Delta_v(T) u_T$, where the coefficient $\Delta_v(T)$ is the Harsanyi dividend~\cite{harsanyi1982simplified} of coalition $T$ in the game, given as, 

\begin{equation}
   \Delta_v(T)=\sum_{S \subseteq T}(-1)^{|T|-|S|} v(S) 
\end{equation}

\noindent The solution concept of cooperative game theory deals with how to allocate the worth $v(N)$ of grand coalition $N$ among the players. Formally, a payoff vector of a game $v$ is an $n$-dimensional vector $x=\left(x_i\right)_{i \in N} \in \mathbb{R}^N$ allocating a payoff $x_i$ to player $i \in N$.

\begin{definition}
\text { A payoff vector } x \text { for a game }(N, v) \text { satisfies },
\begin{itemize}
    \item \text { \emph{efficiency} if } $\sum_{i \in N} x_i=v(N),$
    
    \item \emph{individual rationality} if the payoff to any player $i$ is at least his own worth in the game, i.e., $x_i \geq v(\{i\})$ for each $i \in N$,
    \item \emph{coalitional rationality} if every coalition $S \subseteq N$ receives from $x$ at least the amount it can obtain by operating on its own, i.e., $x(S)=\sum_{i \in S} x_i \geq v(S)$
\end{itemize}
\end{definition}
The \emph{imputation set} $I(N, v)$ of a game $(N, v)$ is the set of all efficient and individually rational payoff vectors given as, 

\begin{equation}
I(N, v)=\left\{x \in \mathbb{R}^n \mid \sum_{i \in N} x_i=v(N), x_i \geq v(\{i\}) \text { for each } i \in N\right\}
\end{equation}

A \emph{solution} for a coalition game maps all the games to a set of payoff vectors. 

The \emph{core} of a cooperative game as the set of all payoff vectors that fulfill \emph{efficiency} and \emph{coalitional rationality}, for all $(N,v) \in G^N$.

\begin{equation}
    C(N, v)=\left\{x \in \mathbb{R}^N \mid \sum_{i \in N} x_i=v(N)\right. and x(S) \geq v(S), for all \left.S \subseteq N\right\}
\end{equation}

The coalition will not improve by leaving the grand coalition and cooperating on its own if the playoff vector belongs to \emph{core} i.e. the elements of the core are stable payoffs.

\subsection{Coalition Games with Levels Structure}
A collection $\left\{B_1, B_2, \ldots, B_m\right\}$ of subsets of $N$ is called a \textit{partition} on $N$ iff $B_1 \cup \cdots \cup B_m=N$ and $B_k \cap B_l=\emptyset$ for $k \neq l$. A coalition structure on $N$ is a partition of $N$.

A \emph{level structure} on $N$ of degree $h$ is a sequence $\mathscr{B}=\left(B^0, B^1, \ldots, B^h\right)$ of partitions of $N$ with $B^0=\{\{i\} \mid i \in N\}$ such that for each $k \in\{0,1, \ldots, h-1\}$ and $S \in B^k$, there is $A \subseteq B^{k-1}$ such that $S=\bigcup_{T \in A} T$. 

The grand coalition is represented as $B^h=\{N\}$ and the levels are represented as $L S^N . B^k$ is called the $k^{t h}$ level of $\mathscr{B}$ and each $S \in B^k$ is a union of level $k$. For any $T \in B^k$ and $S \subseteq T$, we use the notation $B_S^k$ to represent the union of level $k$ that contains the subset $S$, i.e., $B_S^k=T$. When $S=\{i\}$, we simply represent in $B_{\{i\}}^k$, i.e., $B_i^k$ is the union of level $k$ that includes player $i .$

For any $S=\bigcup_{T \in A} T$ \text { for some } $A \subseteq B$ and $N_B(S)=\{T: T \subseteq S, T \in B\}$ represent the \textit{immediate players} of $S$ w.r.t $B$. We represent the set of \emph{immediate players} of $S \in B^{k+1}$ with respect to the $k^{\text {th }}$ level of $\mathscr{B}$ as $N_k(S)$. For a trivial case in a level structure of $\mathscr{B}$ on $N$ , is represented as $\hat{N}$, if $h=1$.

A triple $(N, v, \mathscr{B})$, where $(N, v) \in G^N$ and $\mathscr{B}$ is a level structure on $N$, represents a cooperative scenario with \emph{level structure},that describe the following cooperative scenario. The players first establish a coalition structure $B^1$ (level one of $\mathscr{B}$ ) as \emph{bargaining groups} for the division of $v(N)$. Next, the coalitions in $B^1$ as players establish themselves again into the coalition structure $B^2$ (the second level of $\mathscr{B}$); and so on until the last level of $\mathscr{B}$ is reached. The set of cooperative games with level structures on $N$ is denoted by $L G^N$. A solution on $N$ for cooperative games with levels structures is a real-valued function $\Psi: L G^N \rightarrow \mathbb{R}^n$. 

For each $k \in\{0,1, \ldots, h\}$, we define a game with levels structure $\left(B^k, v / B^k\right.$, $\left.\mathscr{B} / B^k\right)$ on $B^k$, induced from $(N, v, \mathscr{B})$ by viewing unions of level $k$ as individual players. Indeed, $B^k$ is the set of players at level $k$ in the $k$-level game $\left(B^k, v / B^k\right)$. The worth of a coalition of players of level $k$ in the game $\left(B^k, v / B^k\right)$ is defined as the worth of the subset of all original players that it contains. Formally, the worth of the subset $\left\{S_1, S_2, \ldots, S_t\right\} \subseteq B^k$ of players at level $k$ is defined to be $\left(v / B^k\right)\left(\left\{S_1, S_2, \ldots, S_t\right\}\right)=v\left(S_1 \cup S_2 \ldots \cup S_t\right)$. Generally, we denote a game as $(B, v / B)$ on a partition $B$ of $N$, where for any $A \subseteq B,(v / B)(A)=$ $\bigcup_{S \in A} v(S)$. And, $\mathscr{B} / B^k=\left(B^{k, 0}, B^{k, 1}, \ldots, B^{k, h-k}\right)$ is a levels structure of degree $h-k$ starting with the $k^{\text {th }}$ level of $\mathscr{B}$ given by: for all $r \in\{0,1, \ldots, h-$ $k\}, B^{k, r}=\left\{\left\{U: U \in B^k, U \subseteq U^{\prime}\right\}: U^{\prime} \in B^{k+r}\right\}$. \noindent For example,  $\text { Let } \mathscr{B}=\left(B^0, B^1, B^2, B^3\right) \text { be a levels structure of degree } \\ 3 \text { given by }$  $B^0=\{\{1\},\{2\},\{3\},\{4\},\{5\},\{6\}\}, \sloppy B^1=\{\{1,2\},\{3,4\},\{5,6\}\}, \sloppy B^2=\{\{1,2,3,$, $4\},\{5,6\}\}$, and $B^3=\{\{1,2,3,4,5,6\}\}$.  The levels structure $\mathscr{B}$ by viewing the unions of the first level as individuals is $\mathscr{B} / B^1=\left(B^{1,0}, B^{1,1}, B^{1,2}\right)$ where $B^{1,0}=\{\{\{1,2\}\},\{\{3,4\}\},\{\{5,6\}\}\}, B^{1,1}=$ $\{\{\{1,2\},\{3,4\}\},\{\{5,6\}\}\}$, and $B^{1,2}=\{\{\{1,2\},\{3,4\},\{5,6\}\}\}$.

\noindent The value for cooperative games with levels structure of cooperation is a solution concept in cooperative game theory that is used to determine how the total payoff of a cooperative game should be divided among the players, taking into account the levels or coalitions of players who have agreed to cooperate. In the context of machine learning, the value for cooperative games with levels structure of cooperation can be used as a way of forming intermediate concepts from features for machine learning models. For example, in a classification task, the features of a dataset can be grouped into different levels or coalitions based on their semantic meaning or relevance to the task. The value for cooperative games with levels structure of cooperation can then be used to assign a value or weight to each level, based on its contribution to the accuracy or utility of the machine learning model.

\section{Levels Structure in Real-World Scenarios}
In this section, we introduce five axioms that capture properties expected from coalition of features through levels structure when explaining black-box models. 
Let $N$ be the set of features, and let $L = {L_1, L_2, ..., L_m}$ be the set of levels. For each $i = 1, 2, ..., m$, let $L_i$ be the set of features in level $i$.
\newtheorem{prop}{Definition}
\begin{prop}
(Efficiency). $V(N) = v({L_1, L_2, ..., L_m}) = \sum_{S \subseteq N} v(S)$
\end{prop} 
The value of the grand coalition is equal to the sum of the values of all coalitions.

\begin{prop}
(Additivity). $v(S \cup T) = v(S) + v(T)$ for all disjoint sets $S, T \subseteq N$
The value of the union of two disjoint coalitions is equal to the sum of their individual values.
\end{prop}

\begin{prop}
(Level independence). $v(S) = v(S \cap L_1, S \cap L_2, ..., S \cap L_m)$ for all $S \subseteq N$

\noindent The value of a coalition depends only on the levels of the features in the coalition, and not on the specific values of the features.
\end{prop}

\begin{prop}
(Invariance under level-preserving relabeling). If $\sigma_i$ is a permutation of $L_i$ for each $i = 1, 2, ..., m$, then $v(S) = v(\sigma_1(S \cap L_1), \sigma_2(S \cap L_2), ..., \sigma_m(S \cap L_m))$ for all $S \subseteq N$
If the labels of the features within each level are permuted, but the levels themselves are not changed, then the value of each coalition should remain unchanged.
\end{prop}

\begin{prop}
(Covariance). For any feature $i \in N$, and any constant $c > 0$,
$v(S) = v(S \cup {i}) + c - v(S)$ for all $S \subseteq N \setminus {i}$

\noindent If a feature's value is increased or decreased by a certain amount, then the value of each coalition should be increased or decreased by a proportional amount.
\end{prop}

\noindent Naturally, the question arises, \textit{how do we specify or define these feature coalitions in the context of the explainability of black-box models?} In this work, we employ knowledge engineering to collate features based on specific properties proposed by a knowledge engineer. It is important to note that there is no hard and fast rule or criterion to collate features since the end-user can be interested in explanations produced by different coalitions for the same predictive outcome. As a second example, consider a model that predicts car insurance claims. Then, the possible coalitions that can be engineered by a knowledge engineer (a car insurance expert) could be defined as follows:
\begin{itemize}
    \item \textsf{Customer profile} - gender, age, profession, income, marital status, and years of driving experience.
    \item \textsf{Customer history} - claims made in the past ten years, license status.
    \item \textsf{Vehicle profile} - model, the monetary value of the vehicle, mileage covered, age of the car. 
\end{itemize}
The intermediate concepts can be modelled for individuals and is capable of capturing the probability outcome of the insurance claim of each individual, thus allowing us to represent and collate implicit information in the data and generate explanation for each of the coalition.

\subsection{Levels Structure Applied to Post-hoc Explainability}
Explainability refers to understanding how individual agents make decisions and how their interactions lead to emergent behaviors. Achieving explainability involves using rule-based systems, decision trees, and other techniques to provide insights into agent behavior. Transparency and interpretability of agent actions and outcomes are essential for safe and trustworthy decision-making in MAS applications such as robotics, distributed control, and social simulations\cite{patil2022explainability,patil2022towards,patil2022modelling}.
Contextual Importance and Utility (CIU) is a post-hoc explainability method proposed by Fr\"{a}mling in \cite{Framling_CIU_1995,FramlingThesis_1996}. CIU's origins are in Decision Theory and notably Multiple Criteria Decision Making as decribed in \cite{Framling_EXTRAAMAS2020} and addresses the question of how human preferences can be expressed and modelled mathematically. The \textit{intermediate concepts} presented in \cite{FramlingAISB_1996} correspond to coalitions of features with a level structure. In order to show this, we begin by studying the linear case where the additivity condition $v(S \cup T) = v(S)+v(T)$ holds. In this case, an N-attribute utility function is a weighted sum: 

\begin{equation}
u(x)=u(x_{1},\dots ,x_{N})=\sum _{i=1}^{N}{w_{i}u_{i}(x_{i})},
\label{Eq:n_attribute_utility_function}
\end{equation}

where $u_{i}(x_{i})$ are the utility functions that correspond to the features $x_{1},\dots,x_{N}$. $u_{i}(x_{i})$ are constrained to the range [0,1], as well as $u(x)$ through the positive weights $w_{i}$. 
Since $u_{i}(x_{i})$ are in the range $[0,1]$ and $u(x)$ is also constrained to the range $[0,1]$, the importance of feature $i$ is $w_{i}/\sum _{i=1}^{N}{w_{i}}=w_{i}$ because $\sum _{i=1}^{N}{w_{i}}=1$ by definition. This definition can be extended to the importance of a set of inputs $x_{\{i\}}$ versus another set of inputs $x_{\{I\}}$, where $\{i\}$ and $\{I\}$ are index sets and $\{i\}\subseteq \{I\} \subseteq {1,\dots,N}$. Then we have 

\begin{equation}
    \label{Eq:SetWeight}
    w_{\{i\},\{I\}}=\frac{\sum {w_{\{i\}}}}{\sum {w_{\{I\}}}}.
\end{equation}

This joint importance $w_{\{i\},\{I\}}$ of features in $\{i\}$ relative to features in $\{I\}$ defines how intermediate concepts can group features together into coalitions and level structures for providing higher levels of abstraction in the explanations through tree-like levels structures. 

When studying super-additive cooperative games, a linear model is no longer adequate and the additivity condition no longer holds. This is the case for simple logical functions such as $AND$, $OR$ etc. This is the case of most AI-based systems and notably for machine learning (ML) based models, whose outcomes and results we want to explain and justify. Model-agnostic XAI methods treat such models as black-boxes that define a presumably non-linear model $f(x)$. As there are no known $w_{i}$ values for such functions, CIU defines \textit{Contextual Importance (CI) } as the range of variation $[0,\omega_{i}]$, which we can estimate by observing changes in output values when modifying input values of the features in $\{i\}$ and keeping values of the features $\neg \{i\}$ constant at the ones defined by the instance $x$. This gives us an estimation of the range $[umin_{j,\{i\}}(x),umax_{j,\{i\}}(x)]$, where $j$ is the index of the model output explain. CI is defined as:

\begin{equation}
    \label{Eq:CI_definition}
    \omega_{j,\{i\},\{I\}}(x)=\frac{umax_{j,\{i\}}(x)-umin_{j,\{i\}}(x)}{umax_{j,\{I\}}(x)-umin_{j,\{I\}}(x)},
\end{equation}

where we use the symbol $\omega$ for CI. Equation~\ref{Eq:CI_definition} differs from Equation~\ref{Eq:SetWeight} by being a function of $x$, i.e., the instance to be explained. If the model $f(x)$ is linear, then $\omega_{j,\{i\},\{I\}}(x)$ should be the same for all/any instance $x$. If the model is non-linear, then $\omega_{j,\{i\},\{I\}}(x)$ depends on the instance $x$, which is the reason for calling it \textit{contextual} importance.

The utility values $umin_{j}$ and $umax_{j}$ in Equation~\ref{Eq:CI_definition} have to be mapped to actual output values $y_{j}=f(x)$. If $f$ is a classification model, then the outputs $y_{j}$ are typically estimated probabilities for the corresponding class, so we can consider that $u_{j}(y_{j})=y_{j}$. The general form $u_{j}(y_{j})=Ay_{j}+b$ is suitable also for dealing with most regression tasks. When $u_{j}(y_{j})=Ay_{j}+b$, then CI can be directly calculated as:

\begin{equation} \label{Eq:OriginalCI}
\omega_{j,\{i\},\{I\}}(x)= \frac{ymax_{j,\{i\}}(x)-ymin_{j,\{i\}}(x)}{ ymax_{j,\{I\}}(x)-ymin_{j,\{I\}}(x)}, 
\end{equation}

where $ymin_{j}()$ and $ymax_{j}()$ are the minimal and maximal $y_{j}$ values observed for output $j$. 
However, the joint importance $w_{\{i\},\{I\}}$ can not be greater than one by definition:

\begin{definition}[Joint importance of all features]
\label{Def:JointImpAllFeatures}
The joint importance of all features is one, i.e., when the index set $\{i\}=\{1,\dots,N\}$, then $\omega_{\{i\}} = 1$.
\end{definition}



\textit{Contextual Utility (CU}) corresponds to the factor $u_{i}(x_{i})$ in Equation \ref{Eq:n_attribute_utility_function}. CU expresses to what extent the current values of features in $\{i\}$ contribute to obtaining a high output utility (i.e. payoff) $u_{j,\{i\}}$. CU is defined as

\begin{equation}
CU_{j,\{i\}}(x)=\frac{u_{j}(x)-umin_{j,\{i\}}(x)}{umax_{j,\{i\}}(x)-umin_{j,\{i\}}(x)}. 
\label{Eq:CU}
\end{equation}

When $u_{j}(y_{j})=Ay_{j}+b$, this can be written as:  

\begin{equation}\label{Eq:CU_y}
CU_{j,\{i\}}(x)=\left|\frac{ y_{j}(x)-yumin_{j,\{i\}}(x)}{ymax_{j,\{i\}}(x)-ymin_{j,\{i\}}(x)}\right|, 
\end{equation}

where $yumin=ymin$ if $A$ is positive and $yumin=ymax$ if $A$ is negative. 

\textit{Contextual Influence} can be calculated directly from CI and CU values and expresses how much current values of features in $\{i\}$ influence the output compared to a \textit{baseline} or \textit{reference value} $\phi_{0}$:

\begin{equation}\label{Eq:CIU_InfluenceFinal}
    \phi_{j,\{i\},\{I\}}(x)=\omega_{j,\{i\},\{I\}}(x)(CU_{j,\{i\}}(x) - \phi_{0}).
\end{equation}

It should be emphasized that the baseline $\phi_{0}$ of contextual influence is a utility value, so it can have a constant and semantic meaning, i.e., ``averagely good/bad'', ``averagely typical'', ``best possible'' \etc, that presumably makes sense to humans when used in explanations. In many cases, it intuitively makes sense to use the average utility value $0.5$ as a baseline for all features. However, in many real-life use cases other values would make more sense. For instance, $\phi_{0}=0$ would be the expected value in an intrusion detection system, where every threat detection feature could only have positive influence.  

\section{Experiments}
The experimental evaluation of using levels structures / intermediate concepts for producing post-hoc explanations is done using two data sets: 
\begin{enumerate}
    \item The Titanic data set, which is well-known and frequently used for assessing XAI methods. It has two outcome classes and seven features, of which some are numeric and others are categorical. Two intermediate concepts have been defined by the authors, ``FAMILY'' and ``WEALTH''.
    \item The UCI Cars data set, for which the original authors built a ranking system using rules and intermediate concepts. The original rule set is unknown but a black-box model can be trained on the data set and explained using the original intermediate concepts of the authors. 
\end{enumerate}

\subsection{Titanic Data Set}
\begin{figure}
\centering
\includegraphics[width=0.75\columnwidth]{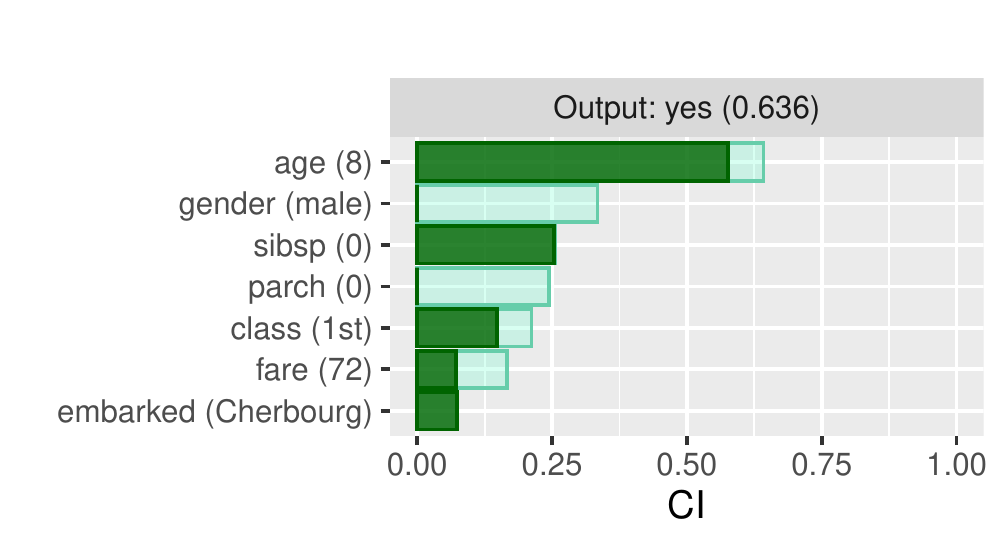}
\caption{CIU bar plot explanation for Titanic, instance `Johnny D'. The transparent bar shows the CI value and the solid bar shows the relative CU value, i.e., how ``favorable'' the current value is.}
\label{Fig:TitanicBarPlotAll}
\end{figure}

The Titanic data set is a classification task with classes `yes' or `no' for the probability of survival. Our Random Forest model achieved 81.1\% classification accuracy on the test set. The training/test set partition was 75\%/25\% of the whole data set. The Titanic data set only has seven features and two output classes but can be considered to be generic, while remaining manageable for presenting the results. The instance `Johnny D' and model are the same as used by Biecek and Burzykowski \cite{BiecekBurzykowski_Book_2021}.
A CIU bar plot explanation is shown in Figure~\ref{Fig:TitanicBarPlotAll} for the probability of survival, which is 63.6\%. 

\begin{figure}[h!]
\centering
\includegraphics[width=0.75\columnwidth]{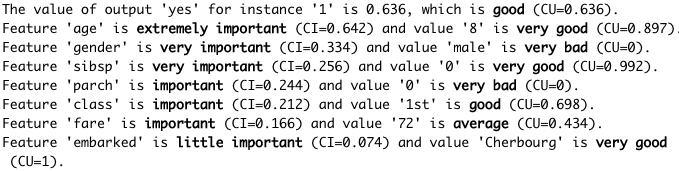}
\caption{Basic textual explanation.}
\label{Fig:TitanicBasicTextual}
\end{figure}

\begin{figure}[!ht]
\centering
\includegraphics[width=0.75\columnwidth]{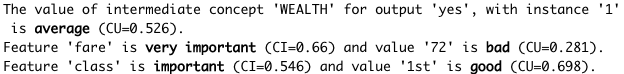} \caption{Explanation for WEALTH intermediate concept.}
\label{Fig:TitanicWEALTH}
\end{figure}

We focus on intermediate concepts WEALTH that combines the input features `fare' (price of ticket) and `class' (cabin class) and FAMILY that combines `sibsp' and `parch'. The levels structure definition in \texttt{R} is:
\begin{lstlisting}[language=R]
wealth<-c(1,6); family<-c(4,5)
gender<-c(2); age<-c(3); embarked <- c(7)
Titanic.voc <- list("WEALTH"=wealth, 
  "FAMILY"=family, "Gender"=gender, 
  "Age"=age, "Embarkment port"=embarked)
\end{lstlisting}
Figures~\ref{Fig:TitanicBasicTextual}-\ref{Fig:TitanicWEALTH} show textual CIU explanations using WEALTH and FAMILY. It is important to point out that \textbf{the CI of an intermediate concept is not simply the sum of its constituent CIs}. Therefore, the CI of FAMILY is not the sum of `parch' and `sibsp' CIs, nor is the CI of WEALTH the sum of `class' and `fare' CIs, unless the underlying model $f$ is linear. 

\begin{figure}

\centering
\includegraphics[width=0.75\columnwidth]{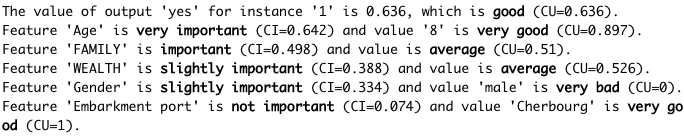}
\caption{Top-level textual explanation with intermediate concepts FAMILY and WEALTH.}
\end{figure}

\begin{figure}

\centering
\includegraphics[width=0.75\columnwidth]{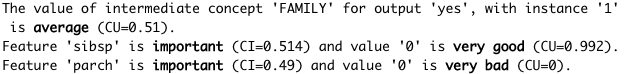} \caption{Explanation for FAMILY intermediate concept.}
\end{figure}

\subsection{UCI Cars Evaluation data set}\label{sec:UCI_Cars}
\begin{figure}
\centering
\includegraphics[width=0.75\columnwidth]{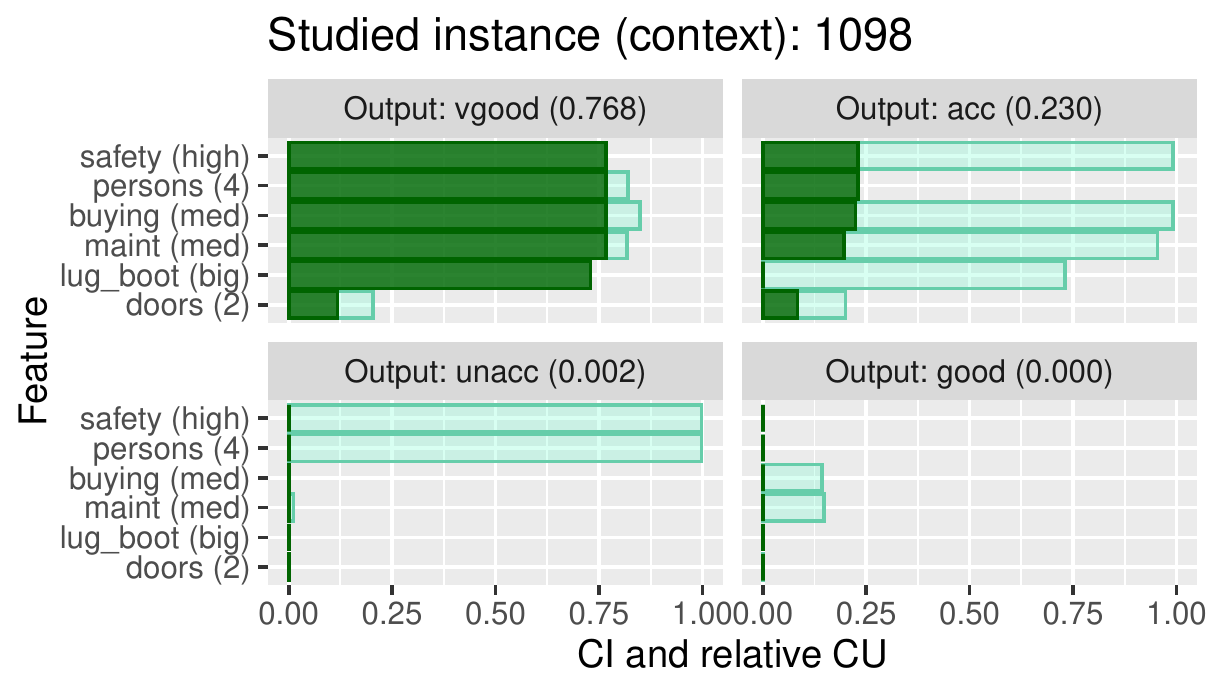}
\caption{CIU bar plot explanation for Car instance \#1098. The transparent bar shows the CI value and the solid bar shows the relative CU value, \ie how ``favorable'' the current value is.}
\label{Fig:Cars}
\end{figure}

The UCI Cars Evaluation data set (\url{https://archive.ics.uci.edu/ml/datasets/car+evaluation}) evaluates how good different cars are based on six categorical features. There are four different output classes: `unacc', `acc', `good' and `vgood'. This signifies that both inputs and output are categorical. Figure \ref{Fig:Cars} shows the basic results for a `vgood' car (instance \#1098). The model is Random Forest. CIU indicates that this car is `vgood' because it has very good values for all important criteria. Having only two doors is less good but it is also a less important feature. In general, the CIU visualisation is well in line with the output value for all classes. The authors of the Cars data set used a rule set with the intermediate concepts `PRICE', `COMFORT' and `TECH', as reported in~\cite{Bohanec88knowledge}. The corresponding levels structure (vocabulary) is defined as follows:

\begin{lstlisting}[language=R]
price <- c(1,2)
comfort <- c(3,4,5)
tech <- c(comfort, 6)
car <- c(price, tech)
voc <- list("PRICE"=price,"COMFORT"=comfort, 
       "TECH"=tech,"CAR"=car)
\end{lstlisting}  

\begin{figure}
\centering
\includegraphics[width=0.75\columnwidth]{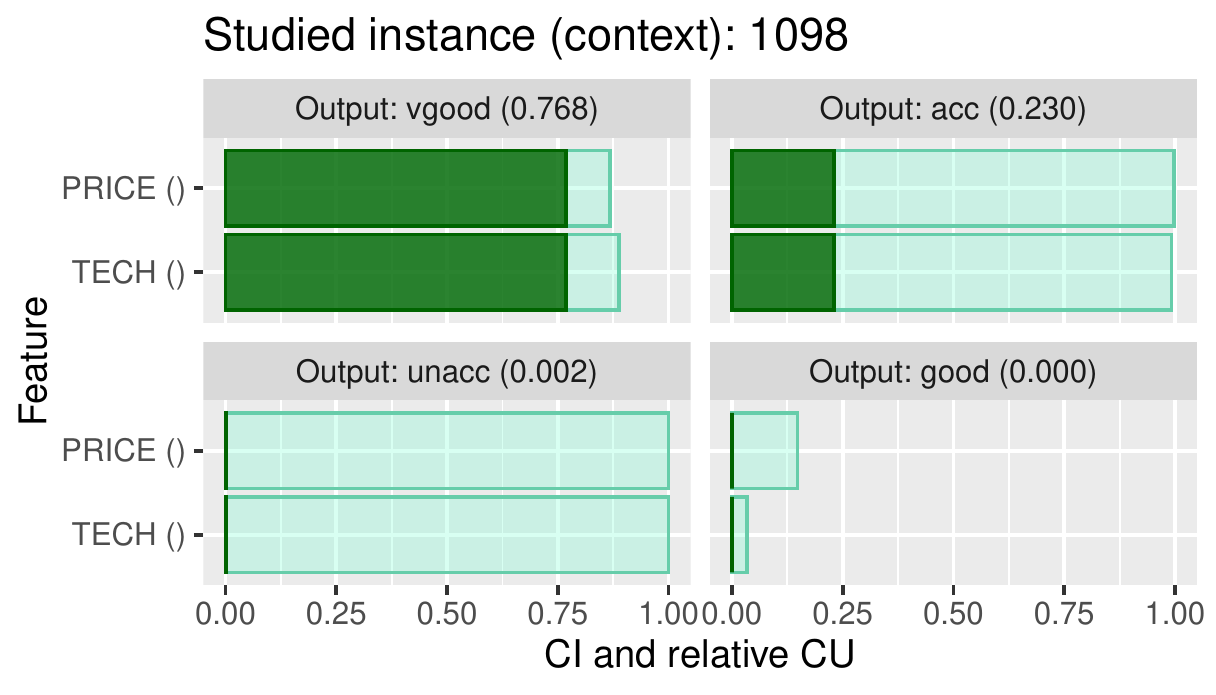}
\caption{Car explanation using highest level intermediate concepts.}
\label{Fig:CarsHighAbstractVocabulary}
\end{figure}

\begin{figure}
\centering
\includegraphics[width=0.75\columnwidth]{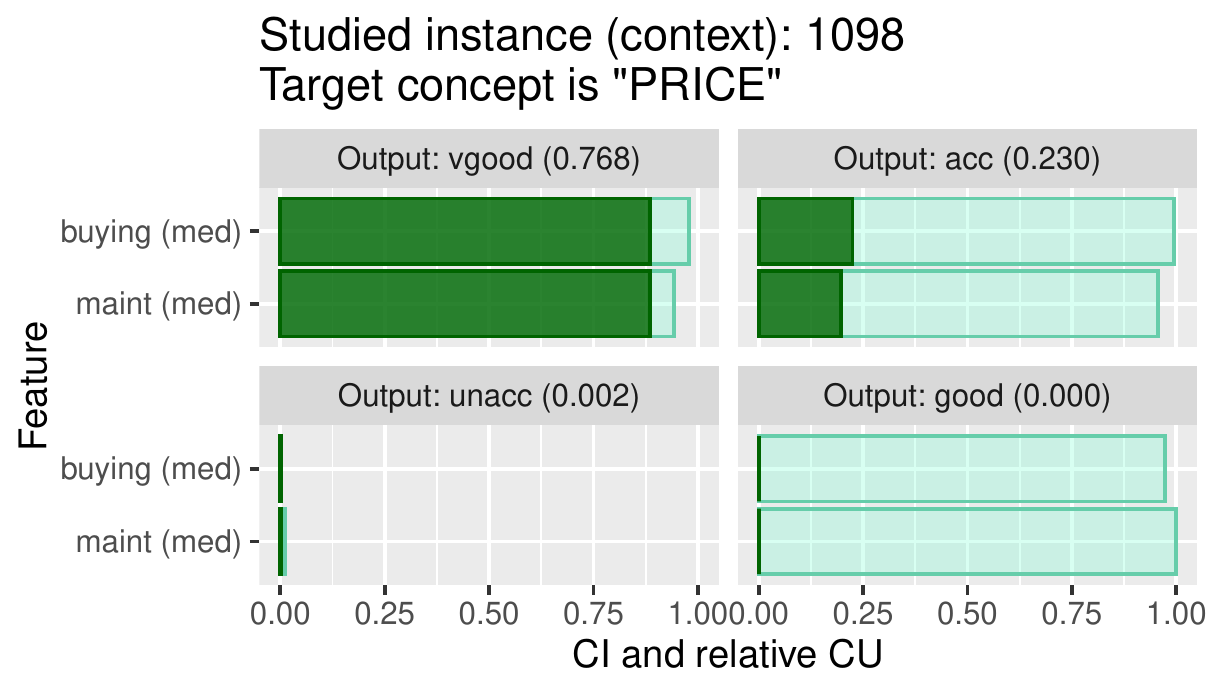}
\includegraphics[width=0.75\columnwidth]{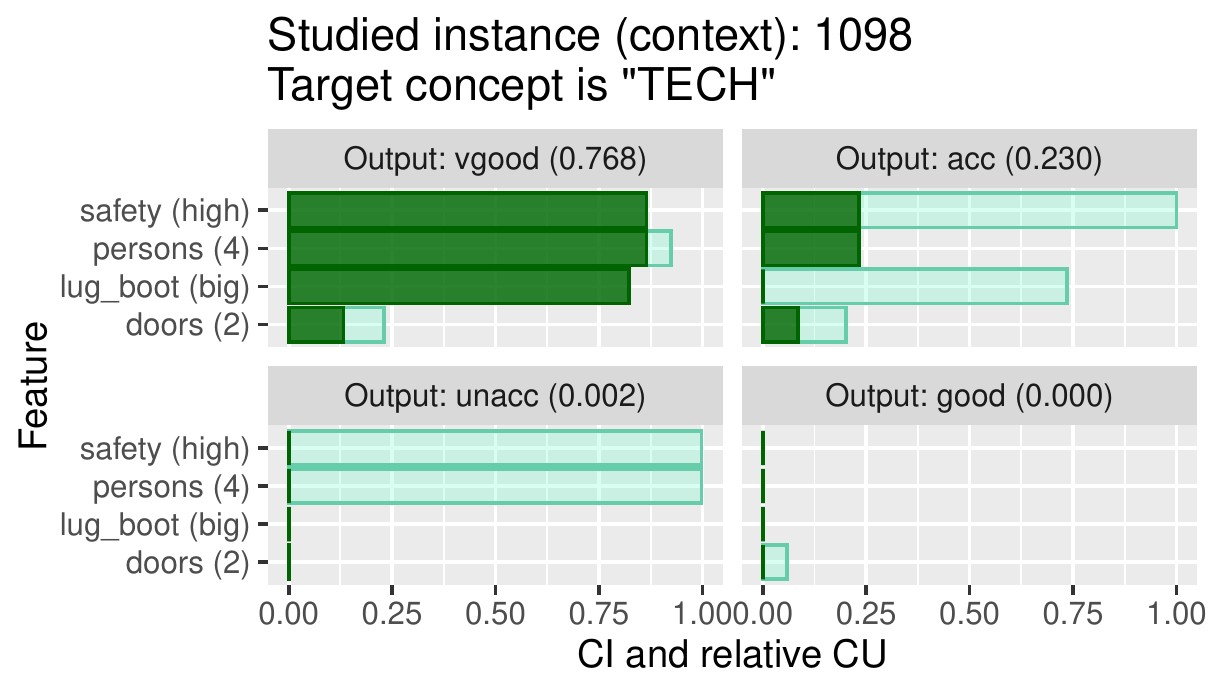}
\caption{Detailed explanations of the two intermediate concepts.}
\label{Fig:CarsVocabulary}
\end{figure}

The corresponding explanations are shown in Figures~\ref{Fig:CarsHighAbstractVocabulary} and \ref{Fig:CarsVocabulary}. 

\section{Discussion}
\begin{figure}[!ht]
\centering
\includegraphics[width=0.75\columnwidth]{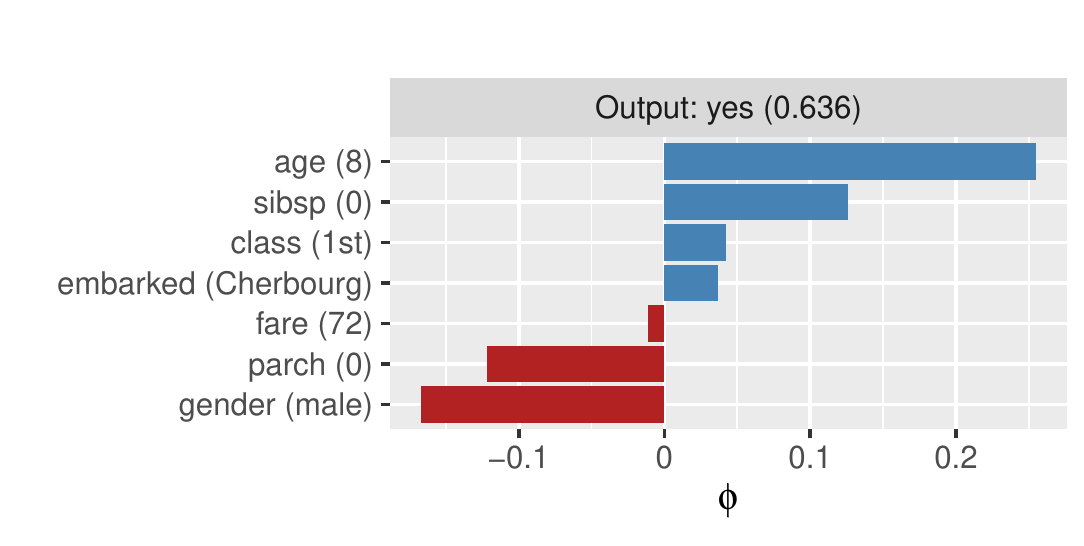}
\caption{Contextual influence bar plot explanation for Titanic, instance `Johnny D'.}
\label{Fig:TitanicCIUinfluence}
\end{figure}

The definition of joint importance in Equation~\ref{Eq:SetWeight} only applies to weights $w$ for linear models and to contextual importance $\omega$. The equations for contextual utility (Equations~\ref{Eq:CU} and \ref{Eq:CU_y}) use the same $umin, umax, ymin, ymax$ values as the CI Equations~\ref{Eq:CI_definition} and \ref{Eq:OriginalCI}, whereas the values $u_{j}(x)$ and $y_{j}(x)$ only depend on the output value for the instance $x$. Contextual influence is also defined for coalitions but we have not studied the use of intermediate concepts and levels structures for them for several reason. To begin with, influence values $\phi$ do not provide the kind of ``counterfactual'' (what-if) explanations as CI and CU in Figure~\ref{Fig:TitanicBarPlotAll}, for instance. Figure~\ref{Fig:TitanicCIUinfluence} shows the corresponding bar plot explanation using Contextual influence. For the FAMILY intermediate concept, we can see that the features `sibsp' and `parch' have opposite signs and would presumably eliminate each other. The same is true for WEALTH, where `class' and `fare' both have small influence values with opposite signs. The true joint Contextual influence values are $\phi_{FAMILY}=0.08$ and $\phi_{WEALTH}=0.04$, so they have close to zero influence on the outcome compared to the used baseline value $\phi_{0}=0.5$. This is correct but it would give the false impression that FAMILY and WEALTH do not have any importance for the outcome, which could be considered misleading. It is indeed a general challenge for influence-based methods that features with average values tend to have small influence values, even though the (contextual) importance of the feature can be significant.  This is also the reason why we have not made any comparisons with the currently most well-known and used post-hoc explainability method, \ie Shapley values \cite{shapley:book1952}. Shapley values are also defined for coalitions. 
As shown in \cite{NIPS2017_Lundberg_XAI}, Shapley values belong to a larger family of XAI methods that they call Additive Feature Attribution (AFA) methods. AFA methods construct an explanation model $g$ that is a linear function of binary variables:
\begin{equation}
g(z')=\phi_{0}+\sum_{i=1}^{M}\phi_{i}z'_{i}, 
\label{Eq:AdditiveFeatureAttribution}
\end{equation}
where $z'\in \{0,1\}^M$, $M\leq N$ is the number of simplified input features, and $\phi \in \mathbb{R}$. For a linear function of the form $f(x)=w_0+w_{1}x_{1}+\dots+w_{N}x_{N}$, the Shapley value $\phi_{i}$ of the $i$-th feature on the prediction $f(x)$ is:
\begin{equation}\label{Eq:Influence}
    \phi_{i}(x)=w_{i}x_{i}-E(w_{i}X_{i})=w_{i}x_{i}-w_{i}E(X_{i}),
\end{equation}
where $E(w_{i}X_{i})$ is the mean effect estimate for feature $i$ \cite{StrumbeljKononenko_2014}.  Equation~\ref{Eq:Influence} is similar to the Contextual influence Equation~\ref{Eq:CIU_InfluenceFinal} when replacing $x_{i}$ with $u_{i}(x_{i})$ as shown in \cite{FramlingAJCAI2022}. Therefore, AFA methods have the same limitations as Contextual influence. 

\begin{figure}[!ht]
\centering
\includegraphics[width=0.75\columnwidth]{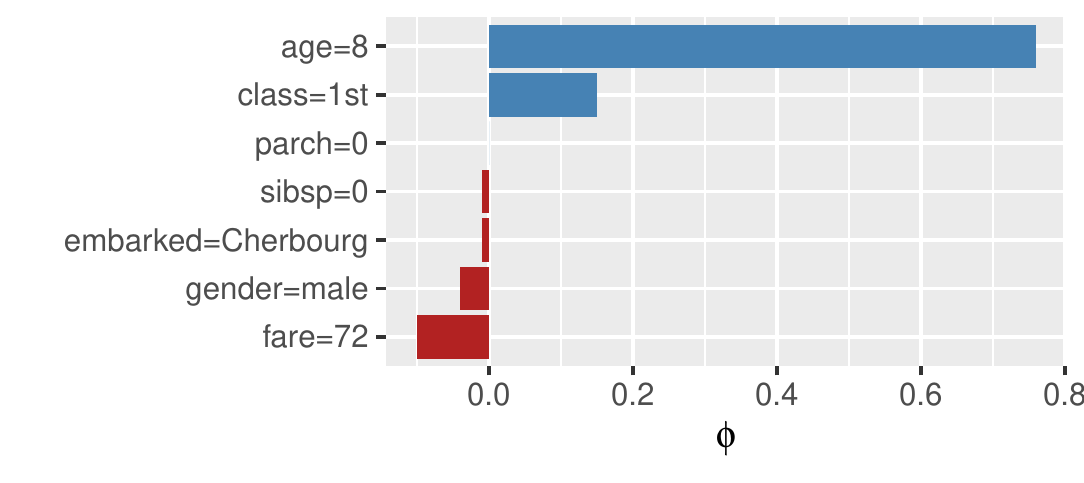}
\caption{Shapley value bar plot explanation for Titanic, instance `Johnny D'.}
\label{Fig:TitanicShapley}
\end{figure}

Figure~\ref{Fig:TitanicShapley} shows the  Shapley values for the Titanic experiment, where it is even clearer than in Figure~\ref{Fig:TitanicCIUinfluence} that features with average values tend to get low influence values. Still, the use of coalitions for Shapley values has been proposed \eg in the PartitionExplainer\footnote{\url{https://shap-lrjball.readthedocs.io/en/latest/generated/shap.PartitionExplainer.html}} and groupShapley approaches \cite{Jullum_GroupShapley_arXiv_2021}. A general challenge is how to deal with dependent features in appropriate ways. Therefore, using level structures for influence values is an ongoing topic and also a topic for future research. 

\section{Conclusion}
The main contributions in our research can be summed up as follows:
\begin{itemize}
    \item Establishing a well-founded link between coalitions in game theory, how they can be used for creating Levels Structures, and how they link with Intermediate Concepts.
    \item Provide the necessary theoretical (mathematical) framework towards Intermediate Concepts as defined for CIU.
    \item Show how such Intermediate Concepts enable the construction of more flexible and expressive explanations than what is possible with the current state-of-the-art post-hoc explainability methods.
\end{itemize}

\end{document}